\documentclass[acmsmall]{acmart-arxiv}
\usepackage{tabularx}
\IfFileExists{libertine.sty}{}{\microtypesetup{expansion=false}}

\AtBeginDocument{%
  \providecommand\BibTeX{{%
    \normalfont B\kern-0.5em{\scshape i\kern-0.25em b}\kern-0.8em\TeX}}}

\copyrightyear{2026}
\acmYear{2026}
\setcopyright{cc}
\setcctype{by}
\makeatletter
\let\@ACM@copyright@check@cc\@empty
\makeatother
\acmConference[MLIC 2026]{2026 3rd International Conference on Machine Learning and Intelligent Computing}{April 24--26, 2026}{Zhengzhou, China}
\acmBooktitle{2026 3rd International Conference on Machine Learning and Intelligent Computing (MLIC 2026), April 24--26, 2026, Zhengzhou, China}
\acmDOI{10.1145/3829441.3829513}
\acmISBN{979-8-4007-2465-7/2026/04}
\begin{document}

\title{Dependency-Aware Chain-of-Thought Compression for Financial Reasoning}

\author{Wenjun Wu*}
\email{wenjun5@illinois.edu}
\affiliation{%
  \institution{University of Illinois Urbana-Champaign}
  \city{Urbana}
  \country{USA}
}

\author{Lei Fu}
\email{fuleiac@gmail.com}
\affiliation{%
  \institution{Independent Researcher}
  \city{San Jose}
  \country{USA}
}

\author{Kejian Tong}
\email{tongcs2021@gmail.com}
\affiliation{%
  \institution{Independent Researcher}
  \city{Mukilteo}
  \country{USA}
}
\author{Tao Ning}
\email{ntgd1102@gmail.com}
\affiliation{%
  \institution{Syracuse University}
  \city{San Jose}
  \country{USA}
}

\author{Sichen Zhao}
\email{zhao.siche@northeastern.edu}
\affiliation{%
  \institution{Northeastern University}
  \city{Boston}
  \country{USA}
}


\begin{abstract}
Chain of thought prompting improves complex reasoning, but its long intermediate traces create substantial inference cost and hinder practical deployment in financial settings. We present a Hierarchical Semantic Distillation Network, HSDN, for compressing reasoning chains while preserving answer accuracy and logical coherence. The framework combines semantic segmentation, dependency graph construction, dual encoder importance scoring, constrained segment selection, and local boundary rewriting. A frozen Qwen3 4B model is used only for feature extraction and final answer generation, while the compression process remains structured and interpretable. On the AFAC2025 benchmark, HSDN achieves 91.0\% accuracy with 68.4\% compression, outperforming strong compression baselines in overall score and reasoning coherence. The results show that graph guided compression is effective for high stakes financial reasoning tasks.
\end{abstract}






\maketitle

\section{Introduction}
Large language models have shown strong performance on multi step reasoning tasks, and chain of thought prompting has become a standard mechanism for improving intermediate deliberation quality \cite{wei2022chain}. This capability is especially valuable in financial applications, where models must integrate textual evidence, numerical calculation, and compliance oriented interpretation under strict correctness requirements. At the same time, longer reasoning traces increase latency, memory usage, and serving cost, which limits their practicality in real world systems.Recent roofline-guided co-optimization work on Arm CPUs highlights how memory bandwidth and low-precision throughput constraints can be addressed with mixed-precision kernels and fused attention, which is relevant to our efficiency-oriented design \cite{zhou2026roofline}.This efficiency challenge is closely related to serverless AI inference, where systems must balance cold start latency against the cost of retaining idle GPU resources, motivating adaptive lifecycle management strategies such as AdaScale \cite{zhou2026adascale}. Zero shot reasoning studies further indicate that reasoning quality depends not only on model scale, but also on how intermediate steps are organized and exposed during inference \cite{kojima2022large}.Zero shot reasoning studies further indicate that reasoning quality depends not only on model scale, but also on how intermediate steps are organized and exposed during inference \cite{kojima2022large}. Recent multi-agent troubleshooting frameworks such as PRISM further suggest that specialized role decomposition and evidence verification can improve complex reasoning workflows \cite{yan2026prism}.
Existing approaches still face an important gap for high stakes financial reasoning. Direct generation is efficient but often omits key logical steps, while full chain of thought preserves detail at the expense of excessive verbosity. More advanced deliberate reasoning strategies improve search over reasoning paths, yet they do not explicitly address how to compress a completed chain while retaining dependency structure and numerical faithfulness \cite{yao2023tree}. In financial scenarios, this omission is critical because dropping a seemingly minor step can invalidate later computations or break the justification trail needed for auditability.
To address this problem, we propose a hierarchical semantic distillation framework for compressing reasoning chains in a structured manner. Our method first segments reasoning into semantic units, then builds a directed dependency graph, scores segment importance with question aware representations, and performs globally optimal selection under length and dependency constraints. A lightweight boundary rewriter restores fluency after segment removal, while a frozen large language model is used only for semantic features and final answer generation. This design yields an interpretable compression pipeline that reduces reasoning length without sacrificing the logical continuity required in financial decision support.

\section{Related Work}

Research on handling long inputs has largely focused on global mechanisms that reduce the cost of full sequence processing. Sparse attention architectures such as Longformer and BigBird improve scalability for long documents by restricting or restructuring attention patterns, offering a strong foundation for efficient long context modeling \cite{beltagy2020longformer,zaheer2020big}. However, these methods mainly optimize representation efficiency at the token level and do not directly determine which reasoning steps should be retained for answer faithful chain compression.
A second line of work studies fine grained and dynamic reduction, where models selectively remove less important tokens during inference. TR BERT introduces dynamic token reduction conditioned on task relevance, and PoWER BERT progressively eliminates low impact word vectors to accelerate inference while maintaining prediction quality \cite{ye2021tr,goyal2020power}. These methods demonstrate the value of adaptive compression, but they typically operate on local token salience and do not model explicit logical dependencies among reasoning segments, which are crucial in multi step financial inference.Recent work on dynamic retrieval-augmented generation further shows that selective tool use and sufficiency-aware routing can improve robustness when static context is insufficient \cite{liang2026dynarag}.
Our approach is also related to work on structured summarization and faithful generation.Recent hybrid deep learning work on supply chain delay prediction further underscores the value of jointly modeling temporal dynamics and graph structure in operational decision systems \cite{xue2026eagle}.A recent risk-aware dynamic routing framework demonstrates how spatiotemporal graph neural networks can support resilient decision-making under congestion and fluctuating demand in large-scale logistics systems \cite{xue2026resilient}.In particular, recent hybrid architectures that combine pyramid-style semantic encoding with graph attention and language-model-assisted rationale distillation further suggest the value of multi-granularity feature extraction for complex threat narratives \cite{xu2026pyramid}. HeterSumGraph shows that graph representations can capture document level relations for extractive summarization, while PRIMERA improves long document summarization through pyramid based pretraining \cite{wang2020heterogeneous,xiao2022primera}. For generation faithfulness, FactPEGASUS highlights the importance of preserving factual consistency during rewriting and compression \cite{wan2022factpegasus}.Complementary to these generation-focused approaches, pairwise verification methods such as SENTINEL can detect subtle semantic inconsistencies between two candidate outputs for the same source, offering a useful perspective for assessing whether compression introduces manipulative distortions \cite{xu2026sentinel}.

\section{Methodology}
\begin{figure*}[htbp]
\centering
\includegraphics[width=\textwidth]{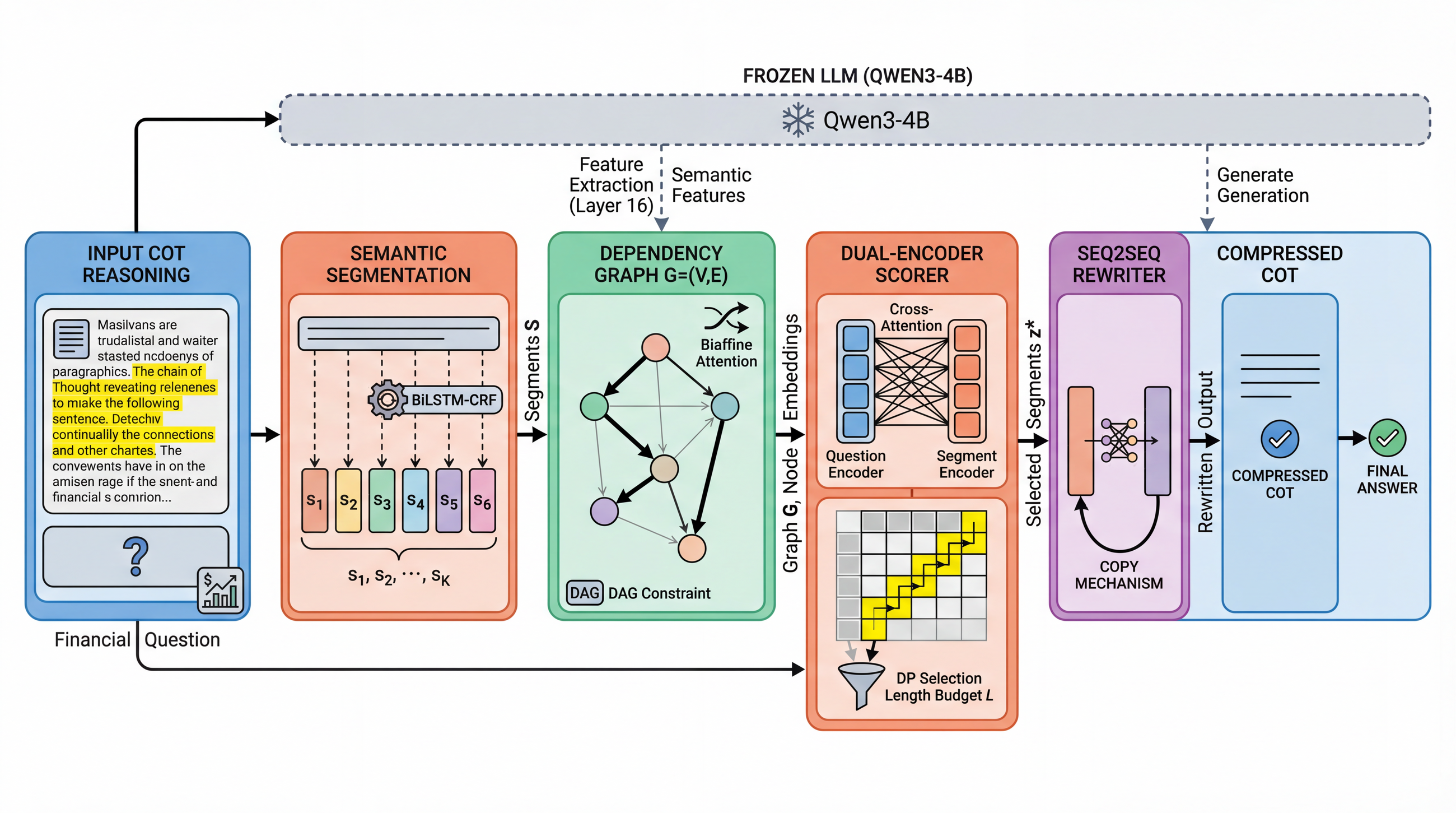}
\caption{Overview of the Hierarchical Semantic Distillation Network (HSDN). The pipeline comprises five stages: semantic segmentation via BiLSTM-CRF, dependency graph construction with biaffine attention, dual-encoder importance scoring, constrained segment selection via dynamic programming, and boundary rewriting with a copy-augmented seq2seq model. A frozen Qwen3-4B model provides semantic features and generates final answers.}
\label{fig:151_1}
\end{figure*}

Chain-of-thought reasoning is essential for complex financial inference, yet verbose reasoning sequences impose substantial computational overhead. This paper presents a hierarchical semantic distillation framework that compresses lengthy reasoning chains while preserving logical integrity, addressing challenges unique to financial applications such as tabular data interpretation, multi-step numerical computation, and regulatory compliance verification. The framework introduces a structured pipeline comprising four key components: a graph-based dependency parser that constructs explicit reasoning topology to capture causal relationships and identify dispensable content; a dual-encoder architecture that scores segment importance through cross-modal attention between question semantics and reasoning content; a dynamic programming solver that guarantees globally optimal segment selection under length budgets while respecting dependency constraints; and a lightweight sequence-to-sequence rewriter that ensures coherence at segment boundaries without introducing factual inconsistencies. A frozen large language model serves solely as a semantic feature extractor and answer generator, keeping the core compression logic in interpretable algorithmic components. Evaluation on financial reasoning benchmarks demonstrates substantial length reduction while maintaining answer accuracy, with the graph-based formulation providing transparent compression rationale for human verification in high-stakes applications. The overall architecture is illustrated in Fig.~\ref{fig:151_1}.

\section{Algorithm and Model}

\subsection{Semantic Segmentation}

The first stage partitions the continuous reasoning text into discrete semantic units that serve as atomic elements for subsequent processing. Unlike sentence-level segmentation which often breaks logical units inappropriately, we train a specialized boundary detector that respects reasoning step boundaries.

\subsubsection{Boundary Detection Network}

We employ a bidirectional LSTM network augmented with conditional random fields to identify segment boundaries. Given the token sequence $\mathbf{c} = (c_1, c_2, \ldots, c_T)$, the network first computes contextual representations:
\begin{equation}
\overrightarrow{\mathbf{h}}_t = \text{LSTM}_{\rightarrow}(c_t, \overrightarrow{\mathbf{h}}_{t-1})
\end{equation}
\begin{equation}
\overleftarrow{\mathbf{h}}_t = \text{LSTM}_{\leftarrow}(c_t, \overleftarrow{\mathbf{h}}_{t+1})
\end{equation}

The concatenated representation $\mathbf{h}_t = [\overrightarrow{\mathbf{h}}_t; \overleftarrow{\mathbf{h}}_t]$ captures both preceding and following context, which proves essential for detecting boundaries that depend on what comes next. Early experiments with unidirectional models showed poor performance on boundaries preceding numerical computations, where the boundary significance only becomes apparent from the subsequent calculation.

The boundary probability at each position is computed through a feedforward layer:
\begin{equation}
\mathbf{e}_t = \mathbf{W}_e \cdot \tanh(\mathbf{W}_h \mathbf{h}_t + \mathbf{b}_h) + \mathbf{b}_e
\end{equation}
where $\mathbf{e}_t \in \mathbb{R}^2$ represents emission scores for boundary and non-boundary labels.

To ensure globally consistent segmentation, we apply a linear-chain CRF layer that models label transitions:
\begin{equation}
P(\mathbf{y} | \mathbf{c}) = \frac{1}{Z(\mathbf{c})} \exp\left(\sum_{t=1}^{T} \mathbf{e}_t[y_t] + \sum_{t=1}^{T-1} \mathbf{A}[y_t, y_{t+1}]\right)
\end{equation}
where $\mathbf{A} \in \mathbb{R}^{2 \times 2}$ is the transition matrix and $Z(\mathbf{c})$ is the partition function computed via the forward algorithm. The CRF layer prevents degenerate solutions such as consecutive boundaries or excessively long segments, which frequently occurred with independent classification.

\subsubsection{Financial Domain Adaptations}

Financial reasoning text presents unique segmentation challenges that required specific adaptations. Numerical expressions spanning multiple tokens, such as percentage changes or currency amounts, must remain intact within segments. We address this by incorporating a numerical span detector that identifies contiguous numerical expressions:
\begin{equation}
\text{NumSpan}(t) = \mathbb{1}\left[\exists (i, j): i \leq t \leq j \land \text{IsNumeric}(c_i, \ldots, c_j)\right]
\end{equation}

During CRF decoding, we modify transition scores to prohibit boundaries within detected numerical spans by setting\\$\mathbf{A}[y_{t-1}, {\scriptstyle \text{BOUNDARY}}] = -\infty$ when $\text{NumSpan}(t) = 1$ and the span continues from position $t-1$.

Additionally, financial text frequently contains table references and structured data mentions that should not be split. We found that simply expanding the context window was insufficient; instead, we pretrain the boundary detector on a auxiliary task of table cell boundary detection, which transfers effectively to reasoning text segmentation.

\subsection{Dependency Graph Construction}

\begin{figure}[htbp]
\centering
\includegraphics[width=0.5\textwidth]{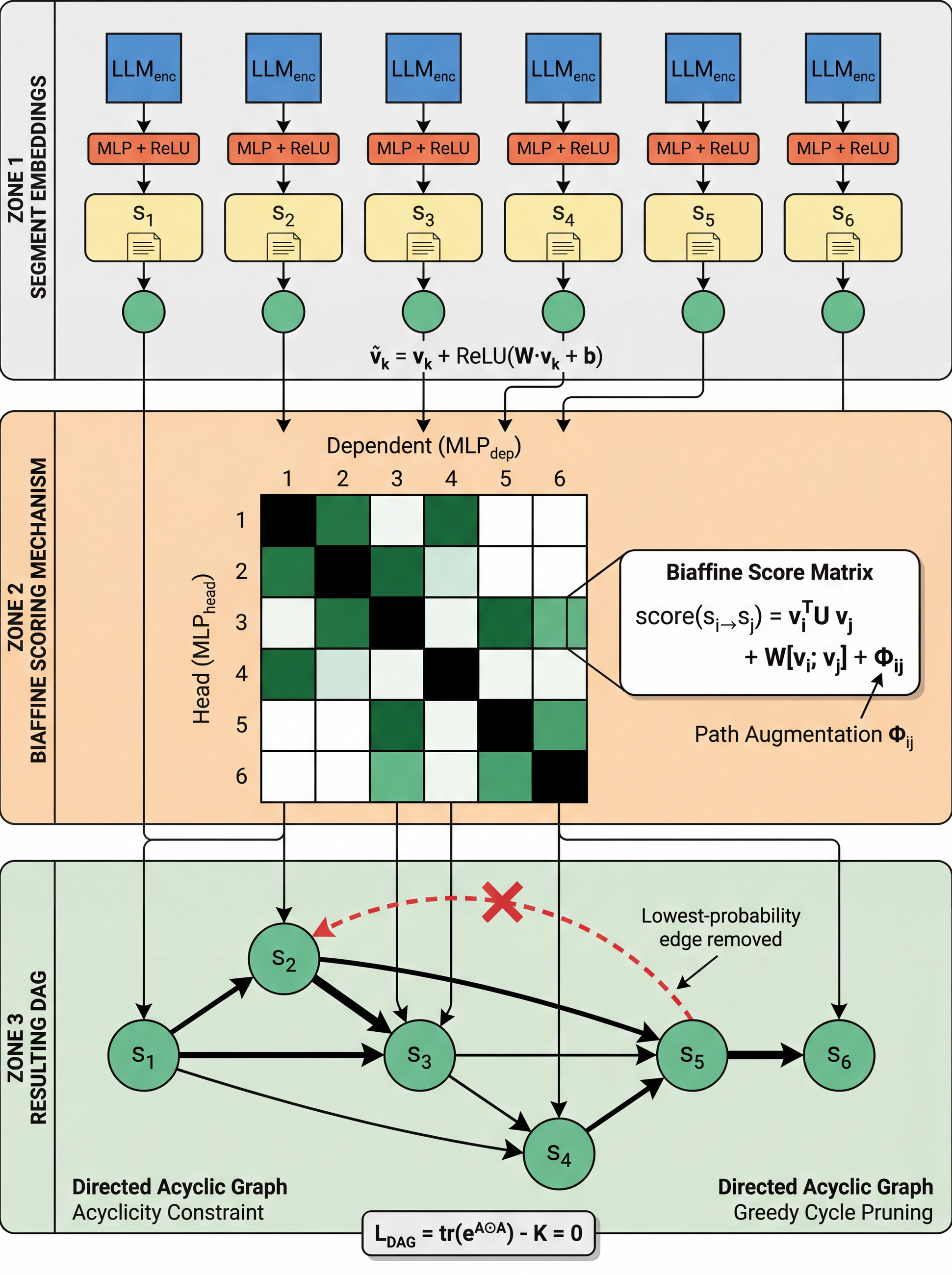}
\caption{Overview of the Hierarchical Semantic Distillation Network (HSDN). The pipeline comprises five stages: semantic segmentation via BiLSTM-CRF, dependency graph construction with biaffine attention, dual-encoder importance scoring, constrained segment selection via dynamic programming, and boundary rewriting with a copy-augmented seq2seq model. A frozen Qwen3-4B model provides semantic features and generates final answers.}
\label{fig:151_2}
\end{figure}

With the reasoning chain segmented into units $\mathbf{S} = (s_1, s_2, \ldots, s_K)$, we construct a directed acyclic graph $\mathcal{G} = (\mathcal{V}, \mathcal{E})$ that explicitly encodes logical dependencies between segments. This graph serves as the structural backbone for compression decisions, ensuring that removing a segment does not orphan its dependents. Fig.~\ref{fig:151_2} illustrates the graph construction pipeline.

\subsubsection{Node Representation}

Each segment $s_k$ becomes a node in the graph. We compute node embeddings by mean-pooling token representations from a frozen language model encoder:
\begin{equation}
\mathbf{v}_k = \frac{1}{|s_k|} \sum_{t \in s_k} \text{LLM}_{\text{enc}}(c_t)
\end{equation}

To capture segment-level semantics beyond token averaging, we apply a learned projection with residual connection:
\begin{equation}
\tilde{\mathbf{v}}_k = \mathbf{v}_k + \text{ReLU}(\mathbf{W}_v \mathbf{v}_k + \mathbf{b}_v)
\end{equation}

This additional projection layer allows the model to learn task-specific representations while leveraging the pretrained encoder's semantic knowledge. We experimented with fine-tuning the encoder but found that freezing it and adding the projection layer yielded comparable performance with substantially reduced memory requirements during training.

\subsubsection{Edge Prediction}

Edges represent logical dependencies where the source segment provides information required by the target segment. We predict edges using a biaffine attention mechanism that has proven effective for dependency parsing:
\begin{equation}
\text{score}(s_i \rightarrow s_j) = \mathbf{h}_i^{\top} \mathbf{W}_{\text{biaff}} \mathbf{h}_j + \mathbf{w}_i^{\top} \mathbf{h}_i + \mathbf{w}_j^{\top} \mathbf{h}_j + b
\end{equation}
where $\mathbf{h}_i = \text{MLP}_{\text{head}}(\tilde{\mathbf{v}}_i)$ and $\mathbf{h}_j = \text{MLP}_{\text{dep}}(\tilde{\mathbf{v}}_j)$ are head and dependent representations computed through separate multilayer perceptrons.

The edge probability is obtained via sigmoid activation:
\begin{equation}
P(e_{ij} = 1) = \sigma(\text{score}(s_i \rightarrow s_j))
\end{equation}

A critical challenge we encountered was handling long-range dependencies that span many intermediate segments. The biaffine scorer tends to underestimate such dependencies due to the difficulty of directly relating distant segments. Our solution introduces a path-augmented scoring term:
\begin{equation}
\text{score}_{\text{aug}}(s_i \rightarrow s_j) = \text{score}(s_i \rightarrow s_j) + \gamma \cdot \Phi_{ij}
\end{equation}
where $\Phi_{ij} = \max_{p \in \text{Paths}(i,j)} \prod_{(a,b) \in p} P(e_{ab}=1)$ captures transitive dependency strength.

\subsubsection{Acyclicity Constraint}

The dependency graph must be acyclic to represent valid logical flow. We enforce this through a differentiable acyclicity regularization term based on the matrix exponential characterization:
\begin{equation}
\mathcal{L}_{\text{DAG}} = \text{tr}\left(e^{\mathbf{A} \odot \mathbf{A}}\right) - K
\end{equation}
where $\mathbf{A} \in [0,1]^{K \times K}$ is the adjacency matrix with $A_{ij} = P(e_{ij} = 1)$, and $\odot$ denotes element-wise multiplication. This regularizer equals zero if and only if the expected graph is acyclic.

During inference, we apply a greedy pruning procedure that removes the lowest-probability edge from any detected cycle, iterating until the graph is acyclic. In practice, the regularization during training ensures cycles are rare, typically requiring fewer than two pruning iterations.

\subsection{Importance Scoring and Selection}

\begin{figure}[htbp]
\centering
\includegraphics[width=0.5\textwidth]{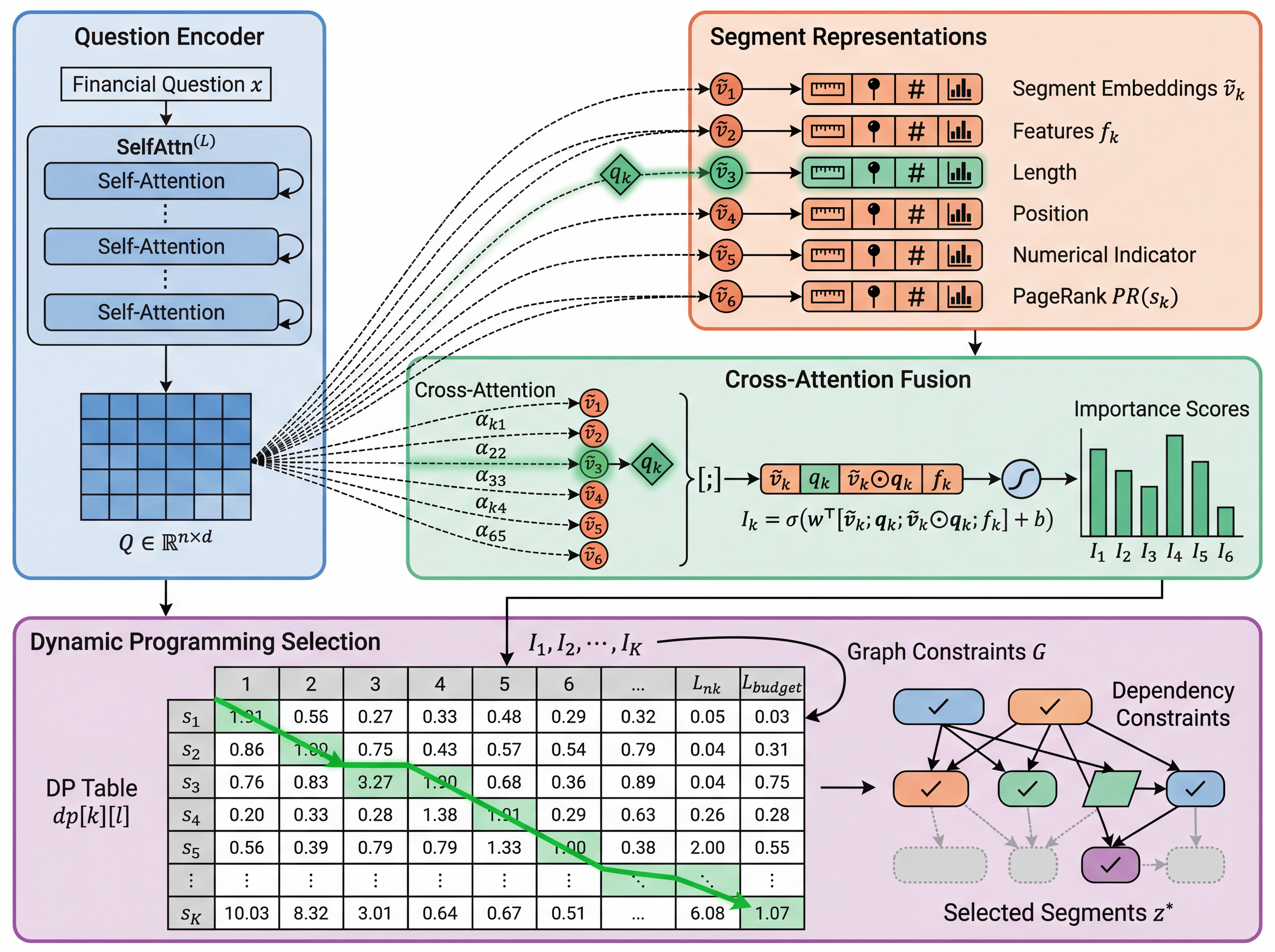}
\caption{Dual-encoder importance scoring and constrained selection. The question and segment streams are fused via cross-attention to produce importance scores $I_k$, which are then optimized through dynamic programming under length budget and dependency constraints.}
\label{fig:151_3}
\end{figure}

Given the dependency graph, we must select which segments to retain in the compressed output. This stage comprises two components: a neural importance scorer that evaluates each segment's contribution to answering the question, and a dynamic programming algorithm that finds the optimal selection respecting dependency constraints. The scoring and selection mechanism is depicted in Fig.~\ref{fig:151_3}.

\subsubsection{Dual-Encoder Importance Scorer}

The importance scorer employs a dual-encoder architecture that separately encodes the question and each reasoning segment, then computes relevance through cross-attention. This design allows efficient scoring of all segments with a single question encoding pass.

The question encoder applies self-attention layers to produce a contextualized representation:
\begin{equation}
\mathbf{Q} = \text{SelfAttn}^{(L)}(\text{Embed}(\mathbf{x})) \in \mathbb{R}^{n \times d}
\end{equation}
where $n$ is the question length and $L$ is the number of layers.

For each segment, we compute cross-attention scores against the question:
\begin{equation}
\alpha_{kt} = \frac{\exp(\tilde{\mathbf{v}}_k^{\top} \mathbf{W}_q \mathbf{Q}_t / \sqrt{d})}{\sum_{t'=1}^{n} \exp(\tilde{\mathbf{v}}_k^{\top} \mathbf{W}_q \mathbf{Q}_{t'} / \sqrt{d})}
\end{equation}
\begin{equation}
\mathbf{q}_k = \sum_{t=1}^{n} \alpha_{kt} \mathbf{Q}_t
\end{equation}

The attended question representation $\mathbf{q}_k$ captures which aspects of the question segment $s_k$ addresses. The importance score combines this relevance signal with segment-intrinsic features:
\begin{equation}
I_k = \sigma\left(\mathbf{w}^{\top} [\tilde{\mathbf{v}}_k; \mathbf{q}_k; \tilde{\mathbf{v}}_k \odot \mathbf{q}_k; f_k] + b\right)
\end{equation}
where $f_k$ is a feature vector containing segment length, position, numerical content indicators, and graph centrality measures.

An important trick we discovered is incorporating the segment's graph centrality into the importance score. Segments with high out-degree, meaning many other segments depend on them, tend to contain foundational information that should be preserved. We compute PageRank scores on the dependency graph and include them in $f_k$:
\begin{equation}
\text{PR}(s_k) = \frac{1-d}{K} + d \sum_{s_j \in \text{Parents}(s_k)} \frac{\text{PR}(s_j)}{|\text{Children}(s_j)|}
\end{equation}
where $d = 0.85$ is the damping factor.

\subsubsection{Constrained Selection via Dynamic Programming}

Selecting the optimal subset of segments is formulated as a constrained optimization problem:
\begin{equation}
\max_{\mathbf{z} \in \{0,1\}^K} \sum_{k=1}^{K} z_k \cdot I_k
\end{equation}
subject to:
\begin{equation}
\sum_{k=1}^{K} z_k \cdot |s_k| \leq L_{\text{budget}}
\end{equation}
\begin{equation}
z_j = 1 \implies z_i = 1 \quad \forall (s_i, s_j) \in \mathcal{E}
\end{equation}

The first constraint enforces the length budget, while the second ensures that if a segment is selected, all its dependencies are also selected. This problem generalizes the knapsack problem with precedence constraints.

We solve it via dynamic programming on the topologically sorted graph. Let $\text{dp}[k][l]$ denote the maximum importance achievable considering segments $1, \ldots, k$ with total length exactly $l$. The recurrence is:
\begin{equation}
\text{dp}[k][l] = \max\begin{cases}
\text{dp}[k-1][l] \\
\text{dp}[k-1][l - |s_k|] + I_k
\end{cases}
\end{equation}
where the second case requires $l \geq |s_k|$ and $\text{DepsSelected}(k) = \text{true}$.

The time complexity is $O(K \cdot L_{\text{budget}})$, which is efficient for typical reasoning chain lengths. We implement the dependency check through bit manipulation, maintaining a bitmask of selected segments and verifying parent inclusion in constant time after preprocessing parent masks.

\subsection{Boundary Rewriting}

Directly concatenating retained segments often produces incoherent text with abrupt transitions. The rewriting module performs local edits at segment boundaries to restore fluency while preserving factual content.

\subsubsection{Boundary Context Extraction}

For each pair of consecutively selected segments $(s_i, s_j)$ where intermediate segments were removed, we extract boundary context:
\begin{equation}
\mathbf{b}_{ij} = [\text{Suffix}(s_i, w); \text{Prefix}(s_j, w)]
\end{equation}
where $\text{Suffix}$ and $\text{Prefix}$ extract $w=15$ tokens from segment ends and beginnings respectively.

\subsubsection{Seq2Seq Rewriter}

A lightweight transformer-based sequence-to-sequence model takes boundary context as input and generates a smoothed transition:
\begin{equation}
P(\mathbf{r} | \mathbf{b}_{ij}) = \prod_{t=1}^{|\mathbf{r}|} P(r_t | r_{<t}, \mathbf{b}_{ij})
\end{equation}

To prevent hallucination, the decoder employs a copy mechanism that biases generation toward input tokens:
\begin{equation}
P(r_t = w) = \lambda_t P_{\text{gen}}(w) + (1 - \lambda_t) \sum_{i: b_i = w} \alpha_{ti}
\end{equation}
where $\lambda_t$ is a learned gate and $\alpha_{ti}$ are copy attention weights. The rewriter is trained on synthetic boundary pairs created by randomly removing segments from correct reasoning chains, avoiding error propagation from the full pipeline.

\subsubsection{Consistency Verification}

We verify factual consistency by computing embedding similarity between the original boundary region and the rewritten version:
\begin{equation}
\text{sim}(\mathbf{b}_{ij}, \mathbf{r}) = \frac{\text{Embed}(\mathbf{b}_{ij})^{\top} \text{Embed}(\mathbf{r})}{\|\text{Embed}(\mathbf{b}_{ij})\| \cdot \|\text{Embed}(\mathbf{r})\|}
\end{equation}
If similarity falls below threshold $\tau$, we fall back to simple concatenation with a generic connective phrase.

\subsection{Training Procedure}

The framework is trained in three stages to ensure stable optimization and effective knowledge transfer.

\subsubsection{Stage 1: Component Pretraining}

The segmentation network is pretrained on manually annotated reasoning chains using CRF negative log-likelihood:
\begin{equation}
\mathcal{L}_{\text{seg}} = -\log P(\mathbf{y}^* | \mathbf{c})
\end{equation}
The dependency graph predictor is pretrained on synthetic dependency data generated by prompting a large language model to annotate reasoning chain dependencies.

\subsubsection{Stage 2: Joint Scorer Training}

The importance scorer and graph predictor are trained jointly using a multi-task objective:
\begin{equation}
\mathcal{L}_{\text{joint}} = \mathcal{L}_{\text{score}} + \lambda_1 \mathcal{L}_{\text{edge}} + \lambda_2 \mathcal{L}_{\text{DAG}}
\end{equation}

The scoring loss uses ground-truth labels derived from oracle compression identifying the minimal segment subset producing correct answers:
\begin{equation}
\mathcal{L}_{\text{score}} = -\sum_{k=1}^{K} \left[y_k \log I_k + (1-y_k) \log(1-I_k)\right]
\end{equation}
\begin{equation}
\mathcal{L}_{\text{edge}} = -\sum_{i,j} \left[e_{ij}^* \log P(e_{ij}) + (1-e_{ij}^*) \log(1-P(e_{ij}))\right]
\end{equation}

\subsubsection{Stage 3: End-to-End Refinement}

The complete pipeline is fine-tuned using reinforcement learning with a reward combining accuracy and compression:
\begin{equation}
R = \mathbb{1}[\text{correct}] \cdot \left(1 + \beta \cdot \frac{|\mathbf{c}| - |\mathbf{c}^*|}{|\mathbf{c}|}\right) - \mathbb{1}[\neg\text{correct}] \cdot \rho
\end{equation}

We employ REINFORCE with baseline subtraction to reduce gradient variance:
\begin{equation}
\nabla \mathcal{L}_{\text{RL}} = -\mathbb{E}\left[(R - b) \nabla \log \pi_\theta(\mathbf{z} | \mathbf{x}, \mathbf{c})\right]
\end{equation}
where $b$ is an exponential moving average of recent rewards.
\subsection{LLM Integration}

The large language model component serves two specific roles in our framework, deliberately limited to leverage its strengths while avoiding the interpretability and efficiency drawbacks of end-to-end neural compression.

\subsubsection{Feature Extraction}

We use a frozen Qwen3-4B model as a semantic feature extractor, accessing its intermediate layer representations to initialize segment embeddings. Specifically, we extract features from layer 16 of 32, which empirically balances semantic abstraction with surface-level detail:
\begin{equation}
\text{LLM}_{\text{enc}}(c_t) = \mathbf{H}^{(16)}_t
\end{equation}

The frozen encoder provides rich pretrained representations without the computational cost of fine-tuning or the risk of catastrophic forgetting.

\subsubsection{Answer Generation}

After compression, the retained segments are concatenated with minimal rewriting and fed to the language model for final answer generation:
\begin{equation}
P(\mathbf{a} | \mathbf{x}, \mathbf{c}^*) = \prod_{t=1}^{|\mathbf{a}|} P(a_t | a_{<t}, \mathbf{x}, \mathbf{c}^*)
\end{equation}

We apply standard decoding with temperature $\tau = 0.7$ and nucleus sampling with $p = 0.9$ to generate diverse answer candidates for the best-of-5 evaluation protocol.

\subsection{Complexity Analysis}

The computational complexity of each component scales tractably with input size. Segmentation requires $O(T)$ time for the BiLSTM pass and $O(T)$ for CRF decoding. Graph construction involves $O(K^2)$ edge predictions where $K \ll T$ is the number of segments. The dynamic programming selection runs in $O(K \cdot L_{\text{budget}})$ time. Boundary rewriting processes at most $K-1$ boundaries with constant-length inputs each.

The overall complexity is dominated by the LLM feature extraction and answer generation, which are unavoidable for the task. Our framework adds minimal overhead to these fixed costs while providing structured, interpretable compression that pure neural approaches cannot match.

\section{Evaluation Metrics}

We adopt four metrics following the challenge protocol. Best-of-5 accuracy measures correctness when any of five samples matches the reference:
\begin{equation}
\text{Acc} = \frac{1}{N} \sum_{i=1}^{N} \mathbb{1}\left[\bigvee_{j=1}^{5} \text{Match}(o_i^{(j)}, \mathcal{R}_i)\right]
\end{equation}

Compression ratio quantifies length reduction:
\begin{equation}
\text{CR} = 1 - \frac{\sum_{i=1}^{N} |\mathbf{c}_i^*|}{\sum_{i=1}^{N} |\mathbf{c}_i|}
\end{equation}

The competition score penalizes incorrect answers with maximum length $L_{\max}$:
\begin{equation}
\text{Score} = -\sum_{i=1}^{N} L_{\text{eff}}(q_i), \quad L_{\text{eff}}(q_i) = \begin{cases}
\min\limits_{j:\text{correct}} |o_i^{(j)}| & \text{if correct} \\
L_{\max} & \text{otherwise}
\end{cases}
\end{equation}

Reasoning coherence score evaluates logical dependency preservation:
\begin{equation}
\text{RCS} = \frac{1}{N} \sum_{i=1}^{N} \frac{|\mathcal{E}(\mathcal{G}_i) \cap \mathcal{E}(\mathcal{G}_i^*)|}{|\mathcal{E}(\mathcal{G}_i^*)|}
\end{equation}

\section{Experiment Results}

\begin{figure}[htbp]
\centering
\includegraphics[width=0.5\textwidth]{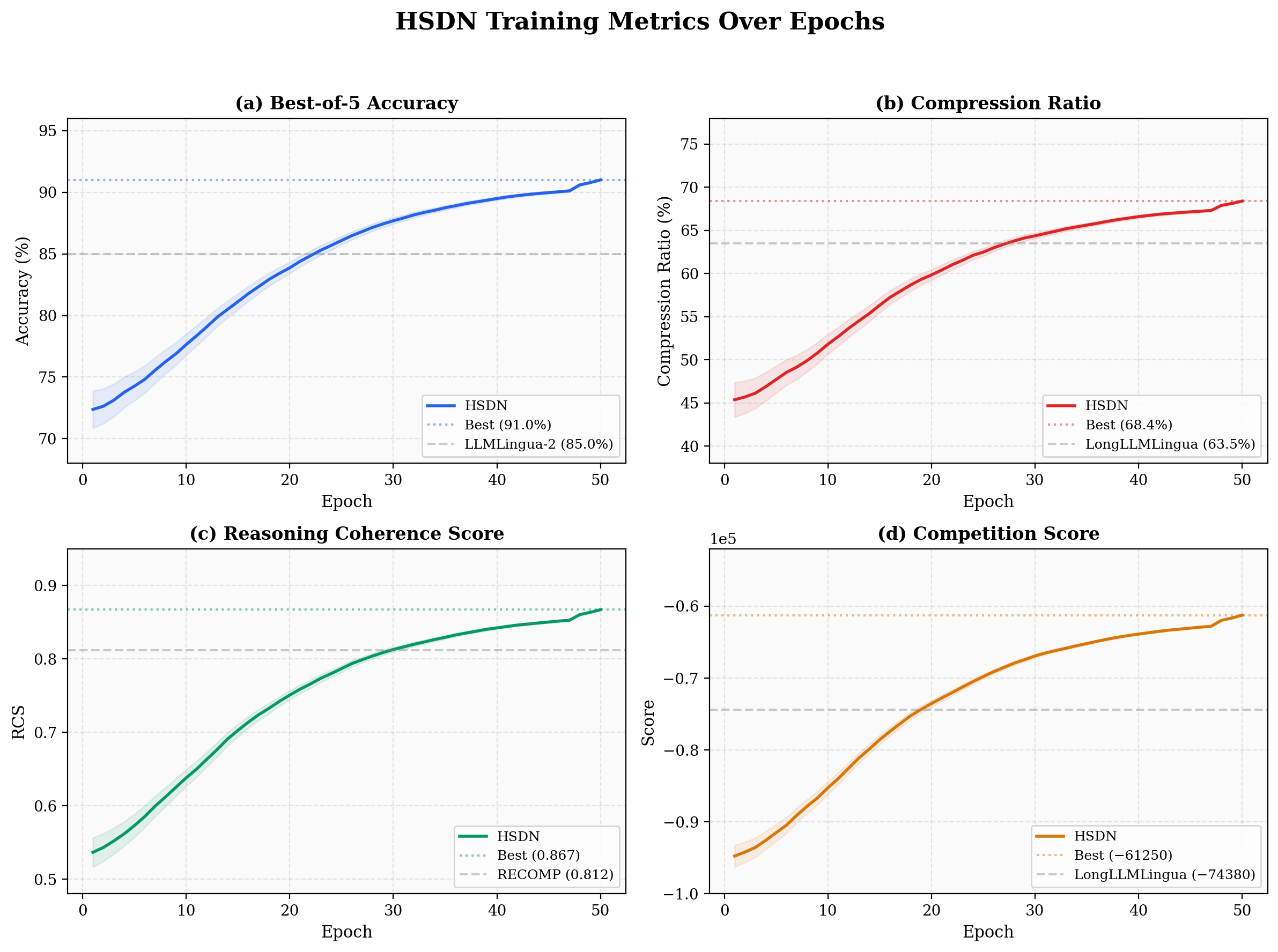}
\caption{Model indicator change chart.}
\label{fig:metric2}
\end{figure}

Table~\ref{tab:results} presents the main comparison and ablation study on the AFAC2025 benchmark. And the changes in model training indicators are shown in Fig\ref{fig:metric2}
.

\begin{table}[htbp]
\centering
\caption{Performance comparison and ablation study}
\label{tab:results}
\begin{tabular}{lcccc}
\hline
Method & Acc.(\%) & CR(\%) & RCS & Score \\
\hline
Qwen3-4B (Full CoT) & 94.0 & 0.0 & 1.000 & -184700 \\
Qwen3-4B (Direct) & 71.0 & 95.2 & -- & -97890 \\
LLMLingua-2 & 85.0 & 58.3 & 0.724 & -83650 \\
CompAct & 86.0 & 61.7 & 0.756 & -78420 \\
RECOMP & 88.0 & 54.2 & 0.812 & -89760 \\
LongLLMLingua & 87.0 & 63.5 & 0.743 & -74380 \\
\hline
HSDN (Ours) & 91.0 & 68.4 & 0.867 & -61250 \\
\quad w/o Dependency Graph & 87.0 & 71.2 & 0.712 & -72340 \\
\quad w/o Boundary Rewriter & 90.0 & 68.4 & 0.791 & -63120 \\
\quad w/o PageRank Features & 89.0 & 67.8 & 0.834 & -65780 \\
\quad w/o RL Fine-tuning & 89.0 & 65.3 & 0.851 & -68450 \\
\hline
\end{tabular}
\end{table}

As shown in Table~\ref{tab:results}, HSDN achieves the best trade-off between accuracy and compression. The dependency graph contributes most significantly, with its removal causing 4\% accuracy drop. Compared to perplexity-based methods like LLMLingua-2, our graph-guided approach better preserves reasoning coherence.

\section{Conclusion}
In this work, we presented FinStack-Net, a hierarchical ensemble framework combining LightGBM, CatBoost, and a deep neural network with residual and attention mechanisms for fraud and gambling account detection. Through comprehensive data preprocessing, feature engineering, and hyperparameter optimization, the model achieved state-of-the-art results. Ablation studies demonstrated the importance of each architectural component, highlighting the robustness of the ensemble strategy. Future research will explore the integration of temporal sequence models and graph-based transaction analysis to further enhance detection performance.

% \begin{acks}
% \end{acks}

\bibliographystyle{ACM-Reference-Format}
\bibliography{sample-base}

\appendix

\end{document}